\documentclass{article}
\usepackage{spconf,amsmath,graphicx, color}
\usepackage{makecell}
\usepackage{hyperref}
\usepackage{amssymb}
\usepackage{mathtools}

\DeclarePairedDelimiter\floor{\lfloor}{\rfloor}

\makeatletter
\def\blfootnote{\xdef\@thefnmark{}\@footnotetext}
\makeatother

\title{Towards efficient Capsule Networks}
\name{Riccardo Renzulli \qquad Marco Grangetto}

\address{Computer Science Department, University of Turin, 10149 Turin, TO, Italy}
\begin{document}
\def\reta{{\textnormal{$\eta$}}}
\def\ra{{\textnormal{a}}}
\def\rb{{\textnormal{b}}}
\def\rc{{\textnormal{c}}}
\def\rd{{\textnormal{d}}}
\def\re{{\textnormal{e}}}
\def\rf{{\textnormal{f}}}
\def\rg{{\textnormal{g}}}
\def\rh{{\textnormal{h}}}
\def\ri{{\textnormal{i}}}
\def\rj{{\textnormal{j}}}
\def\rk{{\textnormal{k}}}
\def\rl{{\textnormal{l}}}
\def\rn{{\textnormal{n}}}
\def\ro{{\textnormal{o}}}
\def\rp{{\textnormal{p}}}
\def\rq{{\textnormal{q}}}
\def\rr{{\textnormal{r}}}
\def\rs{{\textnormal{s}}}
\def\rt{{\textnormal{t}}}
\def\ru{{\textnormal{u}}}
\def\rv{{\textnormal{v}}}
\def\rw{{\textnormal{w}}}
\def\rx{{\textnormal{x}}}
\def\ry{{\textnormal{y}}}
\def\rz{{\textnormal{z}}}

\def\rvepsilon{{\mathbf{\epsilon}}}
\def\rvtheta{{\mathbf{\theta}}}
\def\rva{{\mathbf{a}}}
\def\rvb{{\mathbf{b}}}
\def\rvc{{\mathbf{c}}}
\def\rvd{{\mathbf{d}}}
\def\rve{{\mathbf{e}}}
\def\rvf{{\mathbf{f}}}
\def\rvg{{\mathbf{g}}}
\def\rvh{{\mathbf{h}}}
\def\rvu{{\mathbf{i}}}
\def\rvj{{\mathbf{j}}}
\def\rvk{{\mathbf{k}}}
\def\rvl{{\mathbf{l}}}
\def\rvm{{\mathbf{m}}}
\def\rvn{{\mathbf{n}}}
\def\rvo{{\mathbf{o}}}
\def\rvp{{\mathbf{p}}}
\def\rvq{{\mathbf{q}}}
\def\rvr{{\mathbf{r}}}
\def\rvs{{\mathbf{s}}}
\def\rvt{{\mathbf{t}}}
\def\rvu{{\mathbf{u}}}
\def\rvv{{\mathbf{v}}}
\def\rvw{{\mathbf{w}}}
\def\rvx{{\mathbf{x}}}
\def\rvy{{\mathbf{y}}}
\def\rvz{{\mathbf{z}}}

\def\erva{{\textnormal{a}}}
\def\ervb{{\textnormal{b}}}
\def\ervc{{\textnormal{c}}}
\def\ervd{{\textnormal{d}}}
\def\erve{{\textnormal{e}}}
\def\ervf{{\textnormal{f}}}
\def\ervg{{\textnormal{g}}}
\def\ervh{{\textnormal{h}}}
\def\ervi{{\textnormal{i}}}
\def\ervj{{\textnormal{j}}}
\def\ervk{{\textnormal{k}}}
\def\ervl{{\textnormal{l}}}
\def\ervm{{\textnormal{m}}}
\def\ervn{{\textnormal{n}}}
\def\ervo{{\textnormal{o}}}
\def\ervp{{\textnormal{p}}}
\def\ervq{{\textnormal{q}}}
\def\ervr{{\textnormal{r}}}
\def\ervs{{\textnormal{s}}}
\def\ervt{{\textnormal{t}}}
\def\ervu{{\textnormal{u}}}
\def\ervv{{\textnormal{v}}}
\def\ervw{{\textnormal{w}}}
\def\ervx{{\textnormal{x}}}
\def\ervy{{\textnormal{y}}}
\def\ervz{{\textnormal{z}}}

\def\rmA{{\mathbf{A}}}
\def\rmB{{\mathbf{B}}}
\def\rmC{{\mathbf{C}}}
\def\rmD{{\mathbf{D}}}
\def\rmE{{\mathbf{E}}}
\def\rmF{{\mathbf{F}}}
\def\rmG{{\mathbf{G}}}
\def\rmH{{\mathbf{H}}}
\def\rmI{{\mathbf{I}}}
\def\rmJ{{\mathbf{J}}}
\def\rmK{{\mathbf{K}}}
\def\rmL{{\mathbf{L}}}
\def\rmM{{\mathbf{M}}}
\def\rmN{{\mathbf{N}}}
\def\rmO{{\mathbf{O}}}
\def\rmP{{\mathbf{P}}}
\def\rmQ{{\mathbf{Q}}}
\def\rmR{{\mathbf{R}}}
\def\rmS{{\mathbf{S}}}
\def\rmT{{\mathbf{T}}}
\def\rmU{{\mathbf{U}}}
\def\rmV{{\mathbf{V}}}
\def\rmW{{\mathbf{W}}}
\def\rmX{{\mathbf{X}}}
\def\rmY{{\mathbf{Y}}}
\def\rmZ{{\mathbf{Z}}}

\def\ermA{{\textnormal{A}}}
\def\ermB{{\textnormal{B}}}
\def\ermC{{\textnormal{C}}}
\def\ermD{{\textnormal{D}}}
\def\ermE{{\textnormal{E}}}
\def\ermF{{\textnormal{F}}}
\def\ermG{{\textnormal{G}}}
\def\ermH{{\textnormal{H}}}
\def\ermI{{\textnormal{I}}}
\def\ermJ{{\textnormal{J}}}
\def\ermK{{\textnormal{K}}}
\def\ermL{{\textnormal{L}}}
\def\ermM{{\textnormal{M}}}
\def\ermN{{\textnormal{N}}}
\def\ermO{{\textnormal{O}}}
\def\ermP{{\textnormal{P}}}
\def\ermQ{{\textnormal{Q}}}
\def\ermR{{\textnormal{R}}}
\def\ermS{{\textnormal{S}}}
\def\ermT{{\textnormal{T}}}
\def\ermU{{\textnormal{U}}}
\def\ermV{{\textnormal{V}}}
\def\ermW{{\textnormal{W}}}
\def\ermX{{\textnormal{X}}}
\def\ermY{{\textnormal{Y}}}
\def\ermZ{{\textnormal{Z}}}

\def\vzero{{\boldsymbol{0}}}
\def\vone{{\boldsymbol{1}}}
\def\vmu{{\boldsymbol{\mu}}}
\def\vtheta{{\boldsymbol{\theta}}}
\def\va{{\boldsymbol{a}}}
\def\vb{{\boldsymbol{b}}}
\def\vc{{\boldsymbol{c}}}
\def\vd{{\boldsymbol{d}}}
\def\ve{{\boldsymbol{e}}}
\def\vf{{\boldsymbol{f}}}
\def\vg{{\boldsymbol{g}}}
\def\vh{{\boldsymbol{h}}}
\def\vi{{\boldsymbol{i}}}
\def\vj{{\boldsymbol{j}}}
\def\vk{{\boldsymbol{k}}}
\def\vl{{\boldsymbol{l}}}
\def\vm{{\boldsymbol{m}}}
\def\vn{{\boldsymbol{n}}}
\def\vo{{\boldsymbol{o}}}
\def\vp{{\boldsymbol{p}}}
\def\vq{{\boldsymbol{q}}}
\def\vr{{\boldsymbol{r}}}
\def\vs{{\boldsymbol{s}}}
\def\vt{{\boldsymbol{t}}}
\def\vu{{\boldsymbol{u}}}
\def\vv{{\boldsymbol{v}}}
\def\vw{{\boldsymbol{w}}}
\def\vx{{\boldsymbol{x}}}
\def\vy{{\boldsymbol{y}}}
\def\vz{{\boldsymbol{z}}}

\def\evalpha{{\alpha}}
\def\evbeta{{\beta}}
\def\evepsilon{{\epsilon}}
\def\evlambda{{\lambda}}
\def\evomega{{\omega}}
\def\evmu{{\mu}}
\def\evpsi{{\psi}}
\def\evsigma{{\sigma}}
\def\evtheta{{\theta}}
\def\eva{{a}}
\def\evb{{b}}
\def\evc{{c}}
\def\evd{{d}}
\def\eve{{e}}
\def\evf{{f}}
\def\evg{{g}}
\def\evh{{h}}
\def\evi{{i}}
\def\evj{{j}}
\def\evk{{k}}
\def\evl{{l}}
\def\evm{{m}}
\def\evn{{n}}
\def\evo{{o}}
\def\evp{{p}}
\def\evq{{q}}
\def\evr{{r}}
\def\evs{{s}}
\def\evt{{t}}
\def\evu{{u}}
\def\evv{{v}}
\def\evw{{w}}
\def\evx{{x}}
\def\evy{{y}}
\def\evz{{z}}

\def\mA{{\boldsymbol{A}}}
\def\mB{{\boldsymbol{B}}}
\def\mC{{\boldsymbol{C}}}
\def\mD{{\boldsymbol{D}}}
\def\mE{{\boldsymbol{E}}}
\def\mF{{\boldsymbol{F}}}
\def\mG{{\boldsymbol{G}}}
\def\mH{{\boldsymbol{H}}}
\def\mI{{\boldsymbol{I}}}
\def\mJ{{\boldsymbol{J}}}
\def\mK{{\boldsymbol{K}}}
\def\mL{{\boldsymbol{L}}}
\def\mM{{\boldsymbol{M}}}
\def\mN{{\boldsymbol{N}}}
\def\mO{{\boldsymbol{O}}}
\def\mP{{\boldsymbol{P}}}
\def\mQ{{\boldsymbol{Q}}}
\def\mR{{\boldsymbol{R}}}
\def\mS{{\boldsymbol{S}}}
\def\mT{{\boldsymbol{T}}}
\def\mU{{\boldsymbol{U}}}
\def\mV{{\boldsymbol{V}}}
\def\mW{{\boldsymbol{W}}}
\def\mX{{\boldsymbol{X}}}
\def\mY{{\boldsymbol{Y}}}
\def\mZ{{\boldsymbol{Z}}}
\def\mBeta{{\boldsymbol{\beta}}}
\def\mPhi{{\boldsymbol{\Phi}}}
\def\mLambda{{\boldsymbol{\Lambda}}}
\def\mSigma{{\boldsymbol{\Sigma}}}

\newcommand{\tens}[1]{\boldsymbol{\mathsfit{#1}}}
\def\tA{{\tens{A}}}
\def\tB{{\tens{B}}}
\def\tC{{\tens{C}}}
\def\tD{{\tens{D}}}
\def\tE{{\tens{E}}}
\def\tF{{\tens{F}}}
\def\tG{{\tens{G}}}
\def\tH{{\tens{H}}}
\def\tI{{\tens{I}}}
\def\tJ{{\tens{J}}}
\def\tK{{\tens{K}}}
\def\tL{{\tens{L}}}
\def\tM{{\tens{M}}}
\def\tN{{\tens{N}}}
\def\tO{{\tens{O}}}
\def\tP{{\tens{P}}}
\def\tQ{{\tens{Q}}}
\def\tR{{\tens{R}}}
\def\tS{{\tens{S}}}
\def\tT{{\tens{T}}}
\def\tU{{\tens{U}}}
\def\tV{{\tens{V}}}
\def\tW{{\tens{W}}}
\def\tX{{\tens{X}}}
\def\tY{{\tens{Y}}}
\def\tZ{{\tens{Z}}}

\def\gA{{\mathcal{A}}}
\def\gB{{\mathcal{B}}}
\def\gC{{\mathcal{C}}}
\def\gD{{\mathcal{D}}}
\def\gE{{\mathcal{E}}}
\def\gF{{\mathcal{F}}}
\def\gG{{\mathcal{G}}}
\def\gH{{\mathcal{H}}}
\def\gI{{\mathcal{I}}}
\def\gJ{{\mathcal{J}}}
\def\gK{{\mathcal{K}}}
\def\gL{{\mathcal{L}}}
\def\gM{{\mathcal{M}}}
\def\gN{{\mathcal{N}}}
\def\gO{{\mathcal{O}}}
\def\gP{{\mathcal{P}}}
\def\gQ{{\mathcal{Q}}}
\def\gR{{\mathcal{R}}}
\def\gS{{\mathcal{S}}}
\def\gT{{\mathcal{T}}}
\def\gU{{\mathcal{U}}}
\def\gV{{\mathcal{V}}}
\def\gW{{\mathcal{W}}}
\def\gX{{\mathcal{X}}}
\def\gY{{\mathcal{Y}}}
\def\gZ{{\mathcal{Z}}}

\def\sA{{\mathbb{A}}}
\def\sB{{\mathbb{B}}}
\def\sC{{\mathbb{C}}}
\def\sD{{\mathbb{D}}}
\def\sF{{\mathbb{F}}}
\def\sG{{\mathbb{G}}}
\def\sH{{\mathbb{H}}}
\def\sI{{\mathbb{I}}}
\def\sJ{{\mathbb{J}}}
\def\sK{{\mathbb{K}}}
\def\sL{{\mathbb{L}}}
\def\sM{{\mathbb{M}}}
\def\sN{{\mathbb{N}}}
\def\sO{{\mathbb{O}}}
\def\sP{{\mathbb{P}}}
\def\sQ{{\mathbb{Q}}}
\def\sR{{\mathbb{R}}}
\def\sS{{\mathbb{S}}}
\def\sT{{\mathbb{T}}}
\def\sU{{\mathbb{U}}}
\def\sV{{\mathbb{V}}}
\def\sW{{\mathbb{W}}}
\def\sX{{\mathbb{X}}}
\def\sY{{\mathbb{Y}}}
\def\sZ{{\mathbb{Z}}}

\def\emLambda{{\Lambda}}
\def\emA{{A}}
\def\emB{{B}}
\def\emC{{C}}
\def\emD{{D}}
\def\emE{{E}}
\def\emF{{F}}
\def\emG{{G}}
\def\emH{{H}}
\def\emI{{I}}
\def\emJ{{J}}
\def\emK{{K}}
\def\emL{{L}}
\def\emM{{M}}
\def\emN{{N}}
\def\emO{{O}}
\def\emP{{P}}
\def\emQ{{Q}}
\def\emR{{R}}
\def\emS{{S}}
\def\emT{{T}}
\def\emU{{U}}
\def\emV{{V}}
\def\emW{{W}}
\def\emX{{X}}
\def\emY{{Y}}
\def\emZ{{Z}}
\def\emSigma{{\Sigma}}

\newcommand{\etens}[1]{\mathsfit{#1}}
\def\etLambda{{\etens{\Lambda}}}
\def\etA{{\etens{A}}}
\def\etB{{\etens{B}}}
\def\etC{{\etens{C}}}
\def\etD{{\etens{D}}}
\def\etE{{\etens{E}}}
\def\etF{{\etens{F}}}
\def\etG{{\etens{G}}}
\def\etH{{\etens{H}}}
\def\etI{{\etens{I}}}
\def\etJ{{\etens{J}}}
\def\etK{{\etens{K}}}
\def\etL{{\etens{L}}}
\def\etM{{\etens{M}}}
\def\etN{{\etens{N}}}
\def\etO{{\etens{O}}}
\def\etP{{\etens{P}}}
\def\etQ{{\etens{Q}}}
\def\etR{{\etens{R}}}
\def\etS{{\etens{S}}}
\def\etT{{\etens{T}}}
\def\etU{{\etens{U}}}
\def\etV{{\etens{V}}}
\def\etW{{\etens{W}}}
\def\etX{{\etens{X}}}
\def\etY{{\etens{Y}}}
\def\etZ{{\etens{Z}}}

\newcommand{\enzo}[1]{\textcolor{blue}{Ghost: #1}}
\newcommand{\marco}[1]{\textcolor{orange}{Marco: #1}}
\newcommand{\riccardo}[1]{\textcolor{red}{Riccardo: #1}}
%
\maketitle
\begin{abstract}
From the moment Neural Networks dominated the scene for image processing, the computational complexity needed to solve the targeted tasks skyrocketed: against such an unsustainable trend, many strategies have been developed, ambitiously targeting performance's preservation. Promoting sparse topologies, for example, allows the deployment of deep neural networks models on embedded, resource-constrained devices. 
Recently, Capsule Networks were introduced to enhance explainability of a model, where each capsule is an explicit representation of an object or its parts. 
These models show promising results on toy datasets, but their low scalability prevents deployment on more complex tasks. 
In this work, we explore sparsity besides capsule representations to improve their computational efficiency by reducing the number of capsules. We show how pruning with Capsule Network achieves high generalization with less memory requirements, computational effort, and inference and training time. 
\end{abstract}
%

\section{Introduction}
\label{sec:intro}
\blfootnote{Accepted for publication at the IEEE International Conference on Image Processing (IEEE ICIP 2022).
\\\\
\copyright 2022 IEEE. Personal use of this material is permitted. Permission from IEEE must be obtained for all other uses, in any current or future media, including reprinting/republishing this material for advertising or promotional purposes, creating new collective works, for resale or redistribution to servers or lists, or reuse of any copyrighted component of this work in other works.
}
Capsule Networks (CapsNets)~\cite{hinton-dynamic,hinton-em} were proposed to overcome the shortcomings of Convolutional Neural Networks (CNNs).
For instance, CNNs need many training images in different orientation, pose, size etc. to be robust to affine transformations~\cite{hinton-dynamic}. 
Moreover, CNNs employ max-pooling to both reduce the dimensions of the input space and to route low-level features deeper in the network. However, the spatial relationships between parts (e.g. the mouth or the nose) of an object (e.g. a human face) within an image is lost because max-pooling progressively drops the spatial information~\cite{hinton-dynamic}. 
CapsNets build a hierarchy of parts by means of an iterative routing mechanism~\cite{hinton-dynamic}. This can be seen as a parallel attention mechanism which models the connections between capsules, retaining correct spatial relationships between objects. 

Despite CapsNets are successfully applied to various tasks~\cite{zhao20193d,afshar2018brain}, the research is still in its early stages and CapsNets are far from replacing CNNs as state-of-the-art neural neural architectures. In fact, their applicability is limited because stacking multiple capsule layers increase the computational effort~\cite{light-weight-caps, Q-CapsNets}. Applying CapsNets on high resolution data with complex backgrounds is also critical: stacking many capsule layers or adding more capsules can lead to instability during training because of the delicate routing algorithm~\cite{deepcaps}. Finding the connections between many capsules is not trivial and tuning the number of routing iterations during training can be problematic~\cite{renzulli}.
\begin{figure}
\centering
\includegraphics[width=0.75\columnwidth]{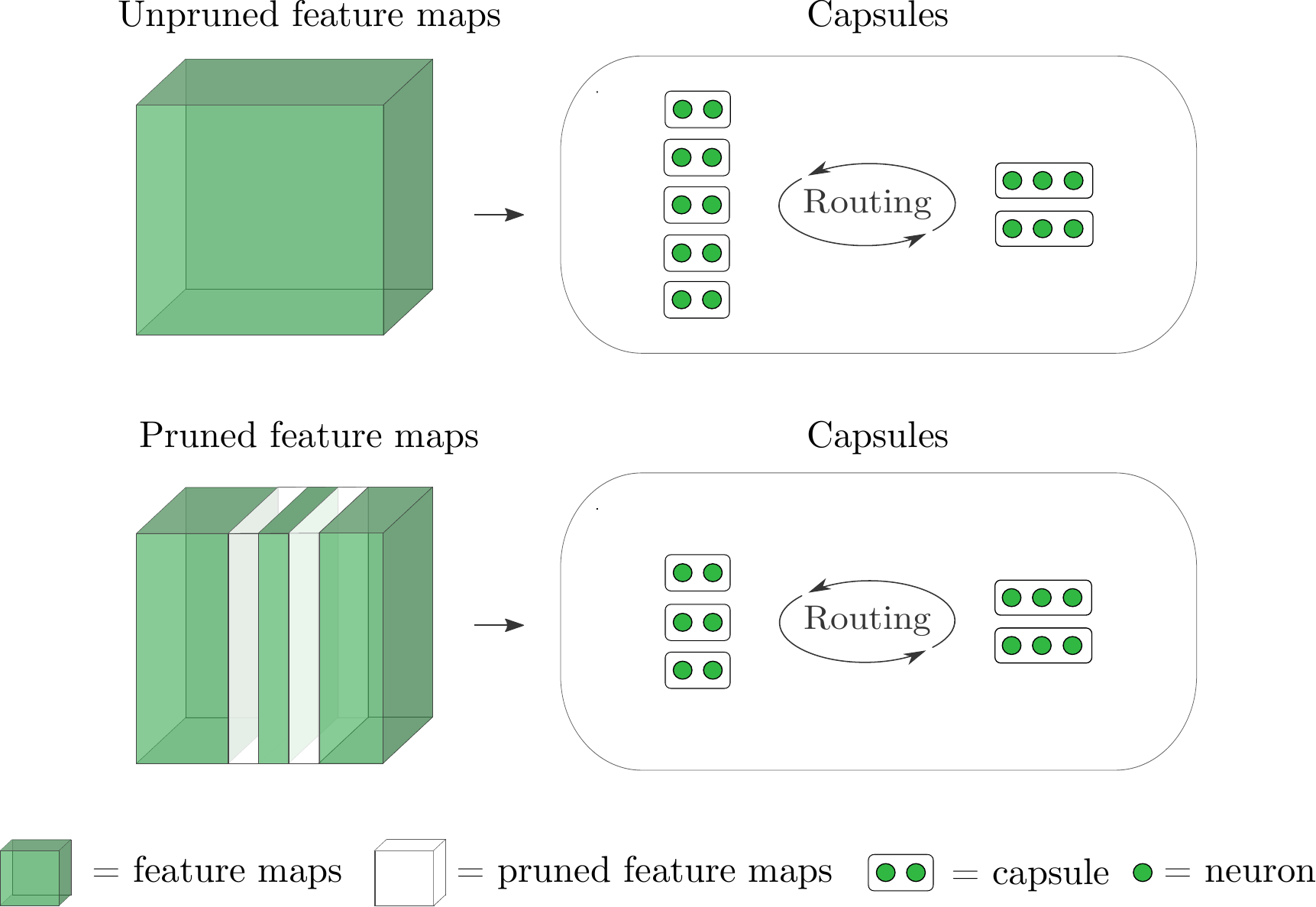}
\caption{A standard CapsNet with many capsules (top). We exploit structured pruned representations (bottom) to reduce the number of capsules.\label{fig:teaser}}
\end{figure}
However, the problem of over-parametrization and the difficulty of deploying networks in scenarios with limited memory and resources is not peculiar only to CapsNets, but also to CNNs. To this end, pruning methods (which reduce the size and complexity of a neural network model) have gained more and more attention lately~\cite{tartaglione2021serene}. Learning sparse topologies can effectively reduce the network footprint and speedup execution with negligible performance loss~\cite{bragagnolo2021simplify}. So, why not exploiting these techniques also for CapsNets? The reduction their complexity, especially in early layers, will enable their deployment on complex datasets.
The aim of this work is to exploit structured
pruning approaches to reduce the complexity of CapsNets without performance loss and test them on data with non-straightforward background such as CIFAR10 and high dimensional data such as Tiny ImageNet. Fig.~\ref{fig:teaser} provides an overall idea of the functioning of our approach.
 We show how employing structurally sparse backbones, extracting features in early layers for CapsNets, affects the number of capsules in the overall network. Our experiments show that reducing the complexity of the backbone is an effective way to achieve high generalization with less memory requirements, energy consumption, training and inference time.


\section{Capsule Networks}\label{sec:sota}
\subsection{Background}
\label{sec:capslayer}
CapsNets organize neurons into activity vectors called \textit{capsules}. Each element of these vectors accounts for the instantiation parameters of the object that the capsule represents such as its pose and other properties like thickness, color, deformation, etc.~\cite{hinton-dynamic}. The probability that a specific object exists in the image is simply the magnitude of its capsule vector~\cite{hinton-dynamic}. CapsNets are organized in layers and are typically composed by convolutional layers and two or more capsule layers. The CapsNet architecture introduced in~\cite{hinton-dynamic} is a shallow network with two capsule layers. In this work, we replace both traditional global average pooling layer and final fully connected layer with these two capsule layers.
 The poses of the first capsule layer (PrimaryCaps)  $\boldsymbol{u}_i$, called \textit{primary capsules}, are built upon convolutional layers reshaped into the capsule space. An iterative \textit{routing-by-agreement} mechanism is performed to compute the poses of the capsules of the next layer (OutputCaps). Each capsule $\boldsymbol{u_i}$ makes a prediction $\hat{\boldsymbol{u}}_{{j|i}}$, thanks to a transformation matrix $\boldsymbol{W}_{ij}$, for the pose of an upper layer output capsule $j$
 
 \begin{equation}
     \hat{\boldsymbol{u}}_{{j|i}} = \boldsymbol{W}_{{ij}}\boldsymbol{u}_{i}.
 \end{equation}

Then, the total input $\vs_{j}$ of an output capsule $j$ is computed as the weighted average of votes $\hat{\vu}_{j|i}$
\begin{equation}\label{eq:sj}
    \vs_{j} = \sum\limits_{i} c_{ij} \hat{\vu}_{j|i},
\end{equation}
where $c_{ij}$ is the coupling coefficient between a primary capsule $i$ and an output capsule $j$.
The un-normalized pose $s_j$ of an output capsule $j$ is then ``squashed'' to compute the normalized pose $\vv_j$ 
 \begin{equation}\label{eq:squashing}
	\vv_{j} = \textrm{squash}(\vs_{j}) = \frac{\|\vs_{j}\|^{2}}{1 + \|\vs_{j}\|^{2}} \frac{\vs_{j}}{\|\vs_{j}\|}.
\end{equation}
The coupling coefficients are determined by a ``routing softmax'' activation function, whose initial logits $b_{ij}$ are the log prior probabilities the $i$-th capsule should be coupled to the $j$-th one.
At each iteration, the update rule for the logits is
\begin{equation}\label{eq:update-rule}
    b_{ij} \gets b_{ij} + \vv_{j} \cdot \hat{\vu}_{j|i}.
\end{equation}
The steps defined in \eqref{eq:sj}, \eqref{eq:squashing}, \eqref{eq:update-rule} are repeated for $r$ routing iterations. Sabour~\emph{et~al.} introduced the \textit{margin loss} to train CapsNets~\cite{hinton-dynamic}.


\subsection{Related work}
Sabour~\emph{et~al.}~\cite{hinton-dynamic} proposed a shallow CapsNet to overcome some limitations of CNNs. They achieved state-of-the-art results on toy datasets like MNIST and smallNORB.
Recent works focused on building deeper and more efficient architectures of CapsNets in order to apply them on more complex data such as CIFAR10. We can divide these attempts into three groups: i) pure CapsNets, namely deep models where only capsule layers are stacked on top of each other~\cite{hinton-em, res-caps, deepcapsnetworkcomplexdata}; ii) CapsNets where the primary capsules are extracted with more than one convolutional layer~\cite{deepcaps, mazzia2021efficient}; iii) CapsNets where the primary capsules are extracted with a pretrained state-of-the-art backbone~\cite{self-routing, paik}. 

As regards pure capsule layers, Hinton~\emph{et~al.}~\cite{hinton-em} replace the dynamic routing with Expectation-Maximization, adopting matrix capsules instead of vector capsules in order to reduce the number of trainable parameters. They also introduce the idea of convolutional capsules, where the routing is performed locally between a subset of capsules defined by a receptive field. Gugglberger~\emph{et~al.}~\cite{res-caps} introduce residual connections between capsule layers to train deeper capsule networks.

A simple but effective way to build deeper capsule models consists of employing several pure convolutional layers before the last capsule layer. Rajasegaran~\emph{et~al.}~\cite{deepcaps} propose a deep CapsNet composed by four layers. All the layers have only one routing iteration (so the coupling coefficients are uniform) except the last one. They approximate the first 3 layers with 2D convolutional layers with skip connections so only one capsule layer exists in their architecture. Mazzia~\emph{et~al.}~\cite{mazzia2021efficient} extract primary capsules by means of a set of convolutional and Batch Normalization layers, replacing dynamic routing with attention routing. They show that even with a very limited number of parameters their network achieves state-of-the-art results on MNIST, MultiMNIST and smallNORB. 

Finally, instead of stacking pure convolutional layers, one can employ state-of-the-arts backbones like ResNets as feature extraction networks.For example, Hahn~\emph{et~al.}~\cite{self-routing} incorporates a  \textit{self-routing} method such that capsule models do not require agreements anymore. In all of their experiments, models share the same base CNN architecture, namely a ResNet-20, requiring less
computation.

Our work relates to those models where primary capsules are extracted using efficient state-of-the-art backbones like ResNets and MobileNets. We employ structurally pruned backbones to scale CapsNets to bigger datasets and improve their efficiency. We show how these pruned Capsule Networks achieve high generalization with less memory requirements and computational effort.
\label{sec:brw}
\section{Proposed method}
\label{sec:method}
\begin{figure*}
\centering
\includegraphics[width=1.5\columnwidth]{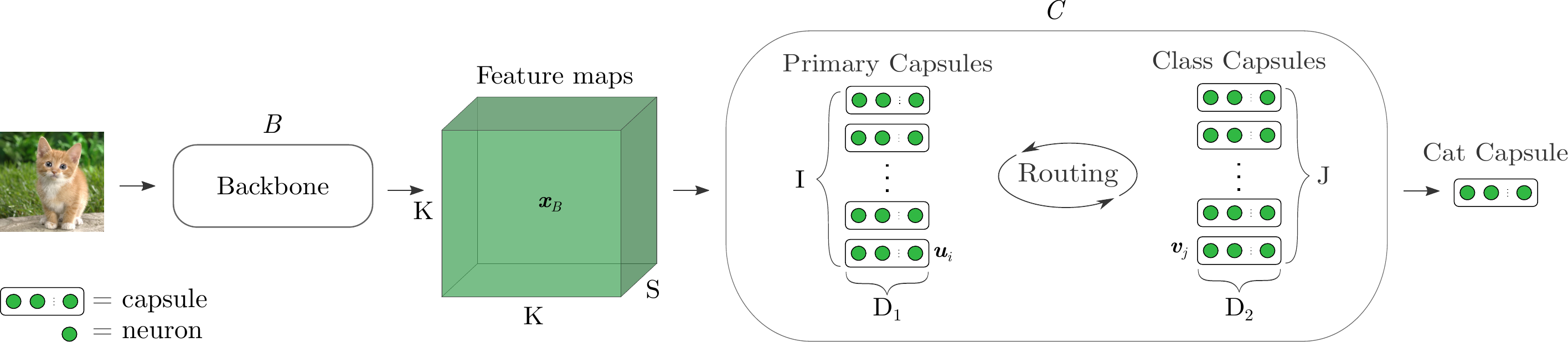}
\caption{The architecture used in this work is composed by a backbone part and a capsule part. Primary capsules space has 8 real dimensions  while output capsules are 16-dimensional vectors ($D_1 = 8$ and $D_2 = 16$).\label{fig:capsnet}}
\end{figure*}
\subsection{Primary capsules extraction}\label{sec:primary_caps_extraction}
Following the work of~\cite{self-routing, paik}, we extract $I$ primary capsule activity vectors by means of a backbone network. With this approach, we can build deep capsule networks in an efficient way since the input space of the capsule network is significantly reduced. Fig.~\ref{fig:capsnet} shows the architecture used in this work. We can build a CapsNet on top of the backbone by replacing the last two layers by a primary capsule (PrimaryCaps) and fully-connected capsule (OutputCaps) layers, respectively. Note that PrimaryCaps is a convolutional capsule layer with $I$ channels of convolutional
$D_1$ capsules, namely each capsule contains $D_1$ convolutional units with a $K \times K \times S$ kernel. 
The kernel size used is wide as the spatial dimensions of the output of the backbone network, which means the number of capsules is dynamically reduced.
In order to compute output capsules, we use the same routing-by-agreement algorithm described in~\cite{hinton-dynamic}.
In our experiments we use backbones both as fixed feature extractor, namely we freeze their weights except for the capsule layer, and as trainable feature extractor, where we finetune all layers.
\subsection{Effect of pruned backbones to capsule layers}
In order to reduce the computational complexity of CapsNets as in Fig.~\ref{fig:capsnet}, one possible strategy is to deploy pruned backbones. Let us define the backbone network $\mathcal{B}(\boldsymbol{x}_\mathcal{M}, \boldsymbol{w}_\mathcal{B})$ where $\boldsymbol{w}_\mathcal{B}$ are the backbone parameters and $\boldsymbol{x}_\mathcal{M}$ is the input; it produces as output the tensor $\boldsymbol{x}_\mathcal{B} \in \mathbb{R}^{K \times K \times S}$. Then, tensor $\boldsymbol{x}_\mathcal{B}$ is fed into the CapsNet $\mathcal{C}(\boldsymbol{x}_\mathcal{B}, \boldsymbol{w}_\mathcal{C})$, where $\boldsymbol{w}_\mathcal{C}$ are the capsule parameters. 
We can concatenate the two parts expressing the end-to-end model $\mathcal{N}$ as 
\begin{equation}
 \mathcal{N}(\boldsymbol{x}_\mathcal{M}, \boldsymbol{w}_\mathcal{B}, \boldsymbol{w}_\mathcal{C}) = \mathcal{C}\left[\mathcal{B}(\boldsymbol{x}_\mathcal{M}, \boldsymbol{w}_\mathcal{B}), \boldsymbol{w}_\mathcal{C})\right].   
\end{equation}
As we saw in Sec.~\ref{sec:primary_caps_extraction}, $D_1$-dimensional primary capsules poses $\boldsymbol{u}_i$ are built upon convolutional layers.
We define the number of primary capsules as 
\begin{equation}\label{eq:num_primary}
    I = \floor*{\frac{S}{D_1}}.
\end{equation}
From this, we observe that reducing the complexity of the backbone would result in the overall reduction of the complexity for the entire model $\mathcal{N}$, and towards this end pruning has already proved to be an effective approach~\cite{tartaglione2021serene, bragagnolo2021simplify}.\\
Pruning approaches can be divided into two groups. \textit{Unstructured} pruning methods aim at minimizing the cardinality $\|\boldsymbol{w}\|_0$ of the parameters in the model, regardless the output topology~\cite{han2015learning, tartaglione2018sensitivity, lobster}. 
On the other hand, \textit{structured} approaches drop groups of weight connections entirely, such as entire channels or filters, imposing a regular pruned topology~\cite{tartaglione2021serene, li2020eagleeye}. As an effect, they minimize the cardinality of some $i$-th intermediate layer's output $\|\boldsymbol{x}_i\|_0$. Bragagnolo~\emph{et~al.}~\cite{bragaICIP} showed that 
structured sparsity, despite removing significantly less parameters from the model, yields lower model's memory footprint and inference time. When pruning a network in a structured way, a simplification step which practically reduces the rank of the matrices is possible; on the other side, encoding unstructured sparse matrices lead to representation overheads~\cite{bragagnolo2021simplify}.\\
Considering the structure of the primary capsules layer, we are interested in the reduction of the cardinality for the input of the primary capsules $\boldsymbol{x}_\mathcal{B}$. Indeed, as we can see in \eqref{eq:num_primary}, the minimization of $\|\boldsymbol{x}_\mathcal{B}\|_0$ results in a proportional reduction of $S$, which in turn reduces $I$, resulting in a reduction of the votes and the complexity of the routing algorithm~\cite{renzulli}.\\

\section{Experiments}\label{sec:experiments}
\begin{table*}[t]
\centering
\small
\begin{tabular}{ccccccc}
\hline
\textbf{\makecell{Backbone\\pruned FLOPS (\%)}} &
\textbf{\makecell{Bottleneck size}} &
\textbf{\makecell{Primary caps}} &
\textbf{\makecell{Total FLOPS (B)}} &
\textbf{\makecell{GPU memory \\consumption (GB)}} & \textbf{\makecell{Training time\\ (epoch, s)}} & \multicolumn{1}{c}{\textbf{Accuracy}} \\ \hline
0 & 2048 & 256 & 12.9 & 47.02          & 462          & 0.78          \\
\textbf{25} & 1970 & 246 &  \textbf{9.8} & \textbf{43.50} & \textbf{450} & \textbf{0.80} \\
\textbf{50} & 1947 & 243 & \textbf{6.7} & \textbf{39.42} & \textbf{410} & \textbf{0.78} \\
75 & 944 & 118 & 3.2 & 20.11          & 214          & 0.74          \\ \hline
\end{tabular}
\caption{\label{tab:performance-dynamiccaps}Dynamic caps Performance per ResNet50 (32 caps, finetuned, Tiny Imagenet).}
\end{table*}
We performed several experiments with different backbones on CIFAR10 and Tiny ImageNet. We choose ResNets and MobileNets networks since they are used as common baselines in traditional computer vision tasks. In fact, they offer a good compromise between both efficiency and performance. Furthermore, they are mainly composed of convolutional layers so it is easy to replace the first convolutional layer in the original CapsNet introduced in \cite{hinton-dynamic} with these models.

\subsection{Experimental Setup}
We resized CIFAR10 images to $64 \times 64$ resolution so as to add a deeper backbone network as in \cite{deepcaps}. We used the 5\% of the training set as validation set for hyper-parameters tuning. We also run experiments on Tiny Imagenet ($224 \times 224$ images, 200 classes). Here we used 10\% of the training set as validation set and the original validation set as test set.
Our experiments consider several flavors of the architecture in Fig.~\ref{fig:capsnet} with different number of primary capsules $I$, backbones (ResNet-50 e MobileNetV1) and training methods (whether the backbone is finetuned or not). 
We train all the networks minimizing a \textit{margin loss}~\cite{hinton-dynamic} with the Adam optimizer and a batch size of 128.  The experiments were run on a Nvidia Ampere A40 equipped with 48GB RAM, and the code uses PyTorch~1.10.\footnote{The source code is available at \url{https://github.com/EIDOSLAB/towards-efficient-capsule-networks}.}\\
We also used PyNVML library to compute the GPU memory
consumption.
As pretrained pruned backbones, we have used both ResNet-50 and MobileNetV1 pruned with EagleEye~\cite{li2020eagleeye}. EagleEye is a state-of-the-art, open-source, structured pruning strategy which identifies relevant features at the level of intermediate outputs $\boldsymbol{x}_i$.

\subsection{Results and discussion}
\begin{figure}
\centering
\includegraphics[width=0.95\columnwidth]{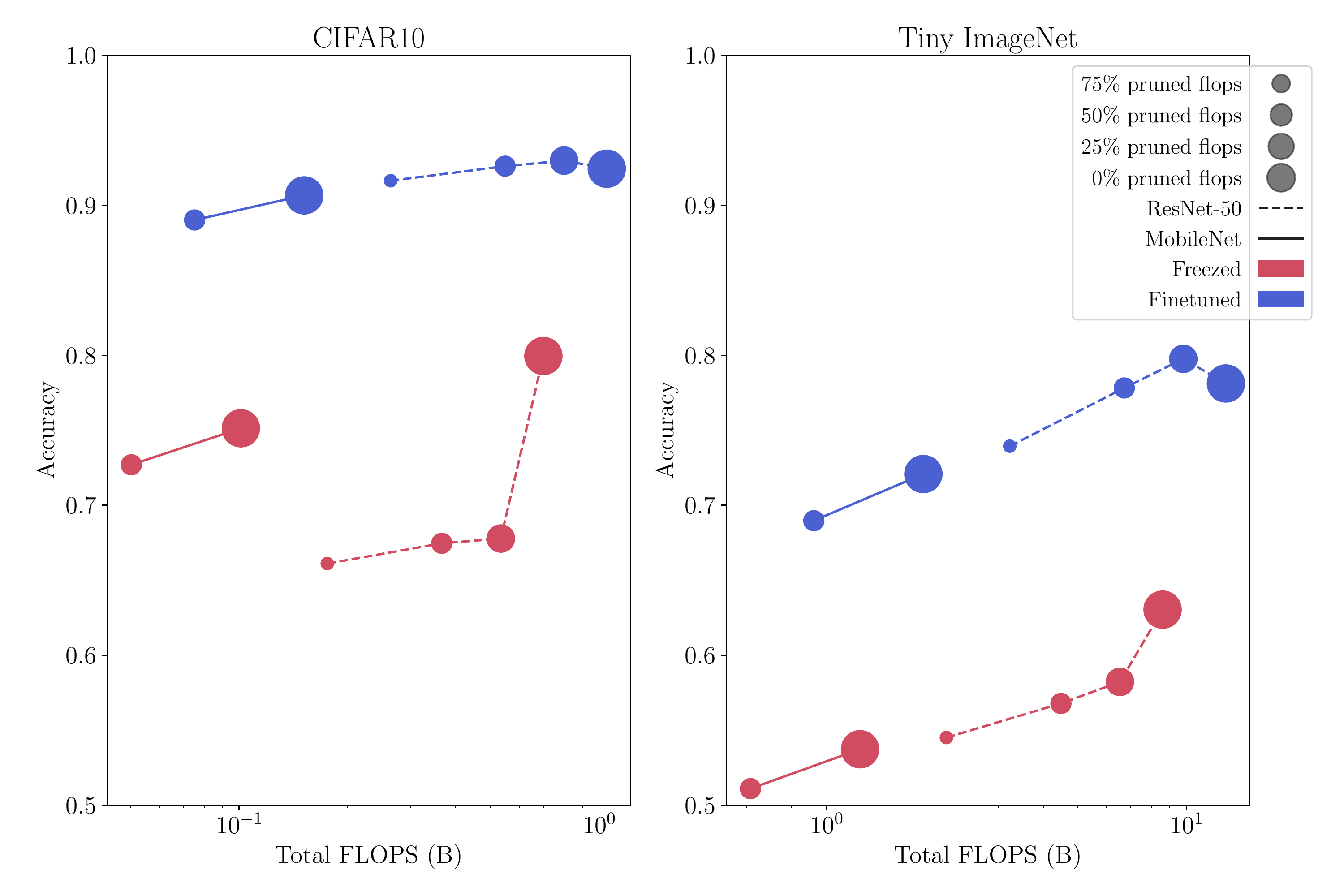}
\caption{Performances of ResNet-50 and MobileNetV1 for CIFAR10 (left) and Tiny ImageNet (right) with several pruned FLOPS. configurations.\label{fig:flops_vs_acc}}
\end{figure}
The main results are shown in Fig.~\ref{fig:flops_vs_acc}. Each configuration differs from the dataset used, percentage of pruned parameters in the backbone network (which affects the number of primary capsules), and training method (backbone finetuned or not). In Fig.~\ref{fig:flops_vs_acc} we report the accuracy as a function of the total number of FLOPS (of both backbone and capsule part) for training the network.
 We can see that when the backbone is freezed (in red) the networks achieve poor performances, but as the number of pruned backbone FLOPS decreases, namely the number of primary capsules is higher, the accuracy increases. The poor performance can be explained by the fact that the output features $\boldsymbol{x}_\mathcal{B}$ of the backbone are not directly optimized to represent objects poses. Therefore, they can not be used directly as input to capsule layers. To overcome such a drawback, we can either add more primary capsules or train the whole network. In fact, when the backbone is finetuned (in blue) the performance is higher since the backbone is able to map more suitable features to the capsules space. We can see in this case that adding more capsules does not always lead to an improvement in the classification accuracy. In Tab.~\ref{tab:performance-dynamiccaps} the performance for Tiny~ImageNet with a pretrained ResNet-50 backbone is reported. Employing a backbone with 50\% of pruned FLOPS yields similar accuracy as with a full backbone, with less GPU memory consumption and training time. With 25\% of pruned FLOPS we can even improve the accuracy compared to the unpruned one. The partial improving in the performance in a low pruning regime, despite being at a first glance surprising, is not new to the literature and it is twofold. First, Rajasegaran~\emph{et~al.}~\cite{deepcaps} shows that building too many capsules inhibites learning because the coupling coefficients are too small, preventing the gradient flow. 
Furthermore, Han~\emph{et~al.} observed such a phenomenon for unstructured pruning~\cite{han2015learning}. Recently, it has been showed that the average entropy in the bottleneck layer for pruned backbones (in our case, $\boldsymbol{x}_\mathcal{B}$) is higher than in non-pruned ones~\cite{bragagnolo2020icann}: this results in the propagation of less specific and more general information, which prevents features overfit, on top of which capsules layers can extract much more accurate information.
\section{Conclusion}
\label{sec:majhead}
In this work we improved the scalability and reduced the computational effort of CapsNets on complex datasets deploying backbones with structured sparsity. 
CapsNets with many capsules are difficult to train: in such a scenario the routing algorithm, a key mechanism for CapsNets, struggles to find the necessary agreements between capsules. Therefore, the employment of  pruned backbones (so fewer capsules) for high resolution datasets leads to competitive results with no performance loss.
We also showed how extracting features in early layers with sparse networks improves efficiency, memory consumption, inference and training time of the overall CapsNet. Therefore, with this work, we open the way to apply CapsNets on resource constrained devices.
Future works involve specific methods concerning pruning capsule layers in a structured way.

\bibliographystyle{IEEE.bst}
\bibliography{main}

\end{document}